\documentclass[letterpaper]{article} 
\usepackage{aaai25}  
\usepackage{times}  
\usepackage{helvet}  
\usepackage{courier}  
\usepackage[hyphens]{url}  
\usepackage{graphicx} 
\usepackage{natbib}  
\usepackage{caption} 
\usepackage{algorithm}
\usepackage{algorithmic}
\usepackage{amsmath}
\usepackage{amssymb}
\usepackage{mathrsfs}
\usepackage{xcolor}
\usepackage{makecell}
\usepackage{multirow}

\usepackage{newfloat}
\usepackage{listings}
\DeclareCaptionStyle{ruled}{labelfont=normalfont,labelsep=colon,strut=off} 
\floatstyle{ruled}
\newfloat{listing}{tb}{lst}{}
\floatname{listing}{Listing}
\title{PromptDet: A Lightweight 3D Object Detection Framework with LiDAR Prompts}
\author{
    Kun Guo,
    Qiang Ling\thanks{Corresponding author.}
}
\affiliations{
    Dept. of Automation, University of Science and Technology of China, Hefei 230027, China

    guok@mail.ustc.edu.cn, qling@ustc.edu.cn
}

\usepackage{bibentry}

\begin{document}

\maketitle

\begin{abstract}
Multi-camera 3D object detection aims to detect and localize objects in 3D space using multiple cameras, which has attracted more attention due to its cost-effectiveness trade-off. However, these methods often struggle with the lack of accurate depth estimation caused by the natural weakness of the camera in ranging. Recently, multi-modal fusion and knowledge distillation methods for 3D object detection have been proposed to solve this problem, which are time-consuming during the training phase and not friendly to memory cost. In light of this, we propose PromptDet, a lightweight yet effective 3D object detection framework motivated by the success of prompt learning in 2D foundation model. Our proposed framework, PromptDet, comprises two integral components: a general camera-based detection module, exemplified by models like BEVDet and BEVDepth, and a LiDAR-assisted prompter. The LiDAR-assisted prompter leverages the LiDAR points as a complementary signal, enriched with a minimal set of additional trainable parameters. 
Notably, our framework is flexible due to our prompt-like design, which can not only be used as a lightweight multi-modal fusion method but also as a camera-only method for 3D object detection during the inference phase. Extensive experiments on nuScenes validate the effectiveness of the proposed PromptDet. As a multi-modal detector, PromptDet improves the mAP and NDS by at most 22.8\% and 21.1\% with fewer than 2\% extra parameters compared with the camera-only baseline. Without LiDAR points, PromptDet still achieves an improvement of at most 2.4\% mAP and 4.0\% NDS with almost no impact on camera detection inference time.
\end{abstract}


\begin{links}
    \link{Code}{https://github.com/lihuashengmax/PromptDet}
\end{links}

\section{Introduction}
\begin{figure}[t]
\centering
\includegraphics[width=1.0\columnwidth]{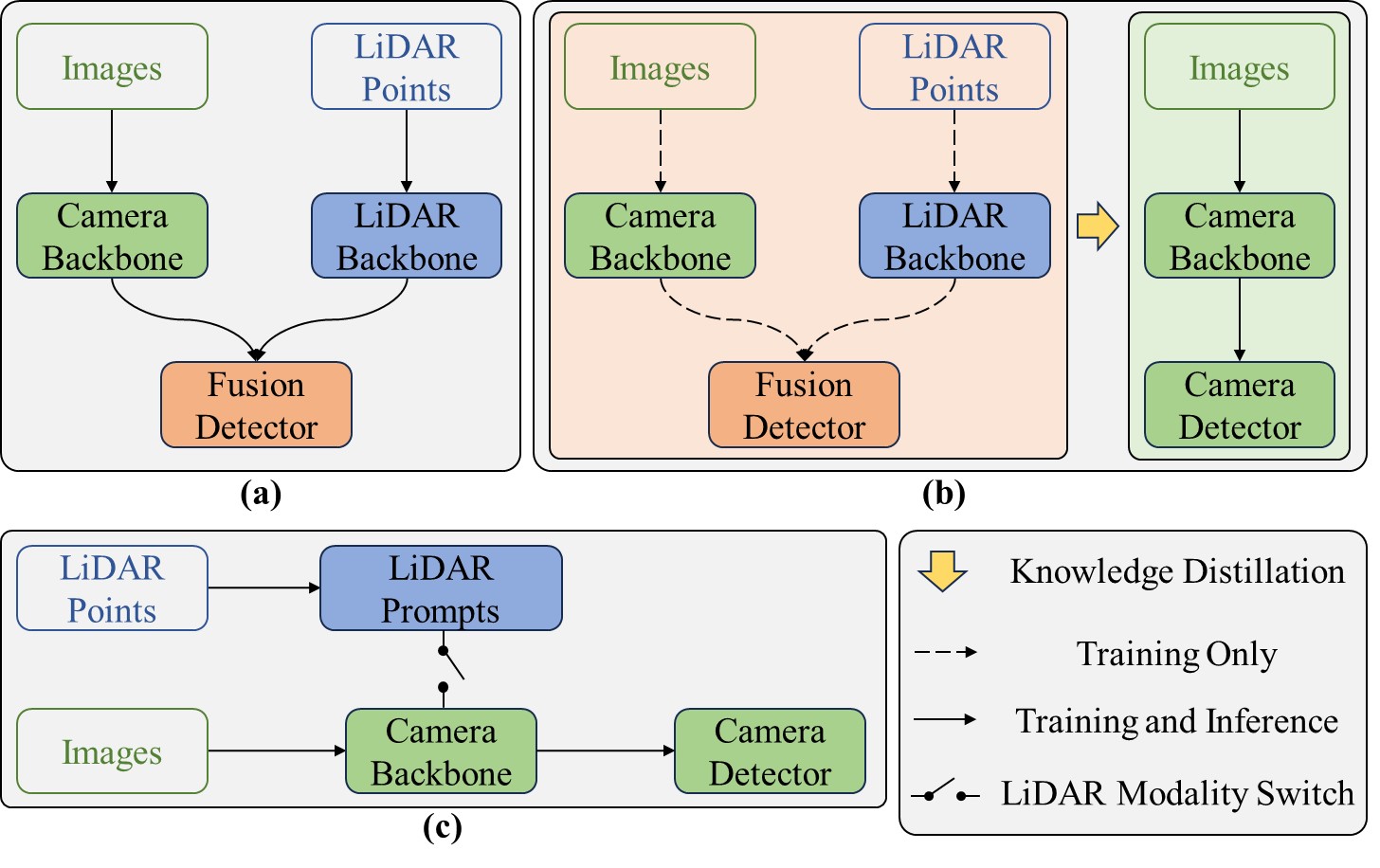}
\caption{Comparison of our PromptDet with previous detection frameworks. (a) Multi-modal detection needs a more complex network architecture. The model training is time-consuming and occupies a huge memory cost. (b) Though knowledge distillation brings performance gains to camera-only detection, a teacher model needs to be trained first and the whole process is laborious and formidable. (c) Our method uses the LiDAR modality as a flexible prompt with a few additional parameters. PromptDet can conduct multi-modal detection and camera-only detection with better performance than the baseline.}
\label{myfig1}
\vspace{-0.5cm} 
\end{figure}
3D object detection is a cornerstone in autonomous driving, with notable advancements in recent years. Most existing camera-based 3D object detectors aim to deduce the spatial arrangement of objects by analyzing visual cues such as color and texture. The low deployment cost and strong scalability on different hardware devices make camera-based detectors quite popular in both academia and industry. Prominent methods like BEVDet \cite{bevdet} and BEVFormer \cite{bevformer} excel in transforming image-based perspective features into a bird's-eye view, showcasing substantial promise for camera-centric 3D detection tasks. However, unlike LiDAR point clouds, the inherent limitation of images is the absence of precise depth data and 3D perceptive capabilities, which poses a challenge to further enhancing the performance of camera-only 3D object detection methods.

To tackle the limitations of camera-only 3D object detection, several multi-modal approaches \cite{bevfusion_beida, bevfusion_mit, cmt, poifusion, msmdfusion, uvtr, zhang2023unleash} have been proposed to leverage the multi-modal data, such as LiDAR points and multi-view images, as depicted in Figure \ref{myfig1} (a). The geometric information provided by LiDAR points is instrumental in offsetting the shortcomings of image-based data. These multi-modal methods achieve better results than the single-modality methods, which demonstrate the effectiveness of the fusion strategies. However, these methods often entail complex network architectures that increase the number of model parameters and computational demands, leading to substantial costs in both the training and inference stages. Moreover, the performance of these multi-modal detectors can significantly degrade in the absence of LiDAR data, potentially falling behind the capabilities of camera-only detectors. Some methods \cite{bevfusion_beida, cmt, metabev, unibev} focus on solving the problem but usually introduce redundant model architecture or utilize mask-modal data augmentation, which brings more training time or data waste. This highlights the need for more efficient and robust approaches that can maintain high performance even in scenarios where certain data modalities are unavailable.

On the other hand, several methods \cite{unidistill, bevdistill, distillbev, simdistill} focus on bolstering camera-only 3D object detectors through the application of knowledge distillation (KD) \cite{kd}, as shown in Figure \ref{myfig1} (b). These methods typically employ a sophisticated, pre-trained multi-modal detector as a ``teacher'' model. This model encapsulates abundant geometric and semantic information about the driving scenario, which is then harnessed to guide and accelerate the learning process of the camera-only ``student'' model. While these KD-based methods have yielded substantial enhancements in the performance of camera-based detectors, the overall training process remains laborious and time-consuming, which necessitates a pre-trained multi-modal detector. This underscores the need for a more efficient training scheme that can effectively impart the necessary knowledge to camera-only 3D object detectors without the excessive overhead associated with traditional KD strategies.

Recently, prompt learning \cite{pl, pl2, pl3, pl4} has drawn increasing attention due to its effectiveness and transferability in both natural language processing (NLP) and computer vision (CV) fields. Intuitively, the paradigm of prompt learning is inherently well-suited for 3D object detection, which can exhibit several advantages, including less memory cost, adaptation to multi-modal input, retaining the base model's potential, and so on. In light of this, we propose PromptDet, a novel lightweight 3D object detection framework, as illustrated in Figure \ref{myfig1} (c). Different from previous works, PromptDet defines a lightweight pipeline, that takes multi-modal detection as injecting LiDAR modality into a camera-based detection model. PromptDet exhibits desirable properties that are not achievable by multi-modal and KD-based methods: (i) The use of LiDAR point clouds is flexible. When LiDAR points are available, PromptDet is a lightweight multi-modal detector with fewer than 2\% extra parameters compared to the camera-only base model but improves performance by at most 22.8\% mAP and 21.1\% NDS. Otherwise, PromptDet degenerates into the original camera-based detector, which can also keep the satisfied performance. (ii) PromptDet can further enhance the performance of the camera-only detector in a single-stage training way. We inject the multi-modal information into the camera-based detector online. With the cross-modal knowledge injection, the camera-based detector learns better feature representations and achieves a performance improvement of at most 2.4\% mAP and 4.0\% NDS.

Specifically, PromptDet consists of a camera-only detector and our proposed plug-and-play LiDAR-assisted prompter, which conducts Adaptive Hierarchical Aggregation(AHA) and Cross-Modal Knowledge Injection(CMKI) to get the complete model. AHA first fuses features from point clouds and images at different scales to get hierarchical multi-modal representations. Then AHA integrates these different-grained features in an adaptive way with several convolutional layers. Built on this design, CMKI makes sure the camera-only feature learns complementary information from the AHA output. In order to guarantee the quality of camera-only and multi-modal features, both of them are supervised by the ground truth with the same network architecture, which we call the hybrid supervision strategy. There are totally only about 1\% additional parameters compared with the camera-only base model during training. 

We summarize the contribution of our work as follows:
\begin{itemize}
    \item We propose a lightweight 3D object detection framework, dubbed PromptDet. The geometry information from point clouds can be injected into the camera-based detector in a flexible way.
    \item We propose adaptive hierarchical aggregation and cross-modal knowledge injection modules, which can transfer the multi-modal information to the camera-based detector efficiently. 
    \item We extensively validate the proposed PromptDet under multi-modal and camera-only settings, which demonstrate the effectiveness of this design.
\end{itemize}

\section{Related Work}
\subsection{Camera-based 3D Object Detection} 
Camera-based 3D object detectors aim to predict the location and class of objects with images only. Numerous recent works \cite{bevdet, bevformer, caddn} find converting perspective features from images to bird's-eye view is effective for camera-based detection. BEVDet \cite{bevdet} explicitly predicts the depth distribution of image features and projects them to the BEV (Bird's Eye View) space through the lifting operation from LSS \cite{lss}. The BEVDet-based approach has been further improved by imposing depth supervision \cite{bevdepth} or temporal aggregation \cite{bevdet4d}, resulting in better performance. BEVFormer \cite{bevformer} samples image features with grid-shaped queries and adds the semantic information to BEV space. FB-BEV \cite{fbbev} proposes a forward-backward projection strategy that generates dense BEV features with strong representation ability through bidirectional projection. Though great progress has been made, camera-based detectors still fall behind LiDAR-based and fusion-based counterparts by a large margin.
\subsection{Knowledge Distillation for Camera Detection}
Camera-based 3D object detection is quite popular due to its low deployment cost and knowledge distillation can improve the detectors' performance while keeping the model architecture unchanged. To fully utilize the information in the 3D world, many researchers employ multi-modal or cross-modal teachers and empower camera-based detectors with more comprehensive representations. DistillBEV \cite{distillbev} proposes an effective balancing design to enable the student to focus on learning crucial features of the teacher with multiple scales. Unidistill \cite{unidistill} is proposed as a universal knowledge distillation framework and supports diverse cross-modal distillation. Simdistill \cite{simdistill} overcomes the cross-modal gap between teacher and student through the simulated LiDAR compensation. These cross-modal distillation techniques endow camera-based detectors with better 3D perceptual capability at the cost of a two-stage complex training process. In a KD-based framework, a teacher model needs to be trained first, which is usually a cross-modal detector or multi-modal detector. Then both the teacher and student are sent to memory and the weights for the teacher model are loaded. Finally, the student model starts to be trained and it receives supervision from both the ground truth and the teacher. The multi-stage training procedure is tiresome and in the final stage, there are two models (student and teacher) loaded into the memory, which brings extra memory cost.
\subsection{Prompt Learning in Computer Vision} 
Recently, as a new paradigm, prompt learning has shown its effectiveness and efficiency in computer vision. It directly utilizes the large pre-trained foundation models (\textit{e.g.}, ViT \cite{vit}) and applies their powerful representations in various downstream tasks via a few additional parameters. AdaptFormer \cite{adaptformer} outperforms fully fine-tuned models on action recognition benchmarks and only adds fewer than 2\% extra parameters to the foundation model. Convpass \cite{convpass} constructs convolutional bypasses as adaptation modules to adapt the large ViT. ViPT \cite{vipt} learns modal-relevant prompts and adapts the frozen pre-trained foundation model to multi-modal tracking tasks. These prompt-learning works all inject task-relevant or modal-relevant information through a few additional trainable parameters while keeping the original base model unchanged. In this paper, we exploit the feasibility of migrating the idea of prompt learning to 3D perception and propose a lightweight 3D object detection framework named PromptDet.
\begin{figure*}[t]
\centering
\includegraphics[width=1.0\textwidth]{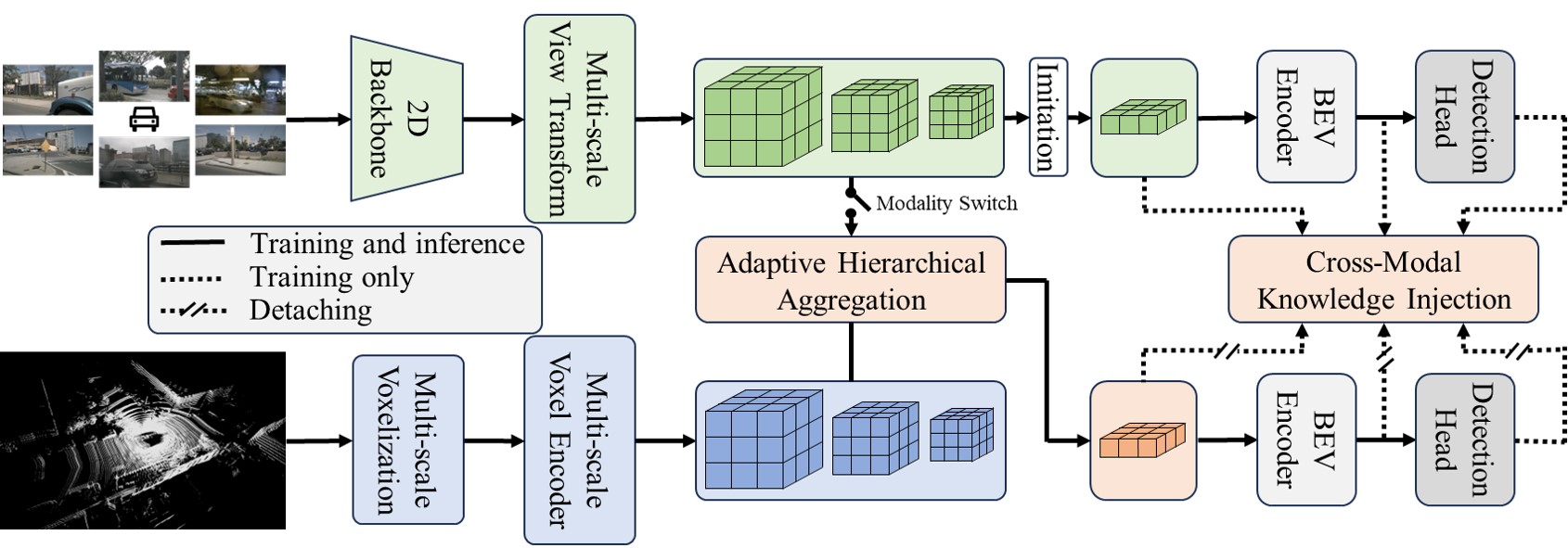}
\caption{The overview of our proposed PrompDet. The model is composed of a camera-only detector and the LiDAR-assisted prompter including Adaptive Hierarchical Aggregation (AHA) and Cross-Modal Knowledge Injection (CMKI). During model training, LiDAR modality switch is turned off. Multi-modal fusion and online knowledge transfer are performed at the same time. PromptDet supports both LiDAR-camera detection and camera-only detection when inference. }
\label{myfig2}
\vspace{-0.5cm} 
\end{figure*}

\section{Methodology}
In this section, we describe our proposed PromptDet in detail. In Section \ref{preliminary}, we introduce the mainstream detection paradigms for LiDAR-only and camera-only 3D object detectors, respectively, and briefly present the knowledge distillation used in Unidistill. In Section \ref{promptDet}, we explain the architecture of PromptDet, which comprises a camera-only detector and the LiDAR-assisted prompter. Finally, in Section \ref{benefits}, we illustrate the superior properties of our method.
\subsection{Preliminary}
Recent LiDAR-only detectors typically convert unstructured point clouds into regular voxels or pillars through voxelization \cite{pointnet, voxelnet, second, pointpillars}, denoted as $F_l$.
These fine-grained features are then extracted and flattened to produce BEV features $F_l^{bev}$. The BEV encoder takes $F_l^{bev}$ as input and generates deeper features $F_l^{en}$. The detection head produces response results $F_l^{resp}$, which include classification and regression heatmaps. Finally, the predictions of the detector are obtained from $F_l^{resp}$.

Inspired by LiDAR-only 3D object detectors, numerous works attempt to transfer the BEV detection paradigm to camera-based detection. The pixel-wise features of perspective images are projected into 3D space through depth estimation and view transformation. Efficient pooling is employed to obtain pseudo-voxel features $F_c$, which are analogous to $F_l$. Following the same procedure as in LiDAR-only detection, $F_c^{bev}$, $F_c^{en}$, $F_c^{resp}$ and the final predictions are generated sequentially.

UniDistill is proposed as a cross-modal knowledge distillation framework to align the foreground BEV features of different modalities and three types of distillation losses are calculated. Denoting the detection modality as $mod$, these losses include feature knowledge distillation on $F_{mod}^{bev}$, relation knowledge distillation on $F_{mod}^{en}$, and response knowledge distillation on $F_{mod}^{resp}$.
\label{preliminary}
\subsection{PromptDet}
Although prompt learning has been proven effective, its direct transfer to 3D object detection is challenging due to the absence of a 3D universal foundation model pre-trained on large-scale datasets. To address this issue, we select the BEVDet-series detectors as the base model and introduce the LiDAR modality as a prompt. The overview of our proposed PromptDet framework is illustrated in Figure \ref{myfig2}.
The LiDAR-assisted prompter consists of Adaptive Hierarchical Aggregation (AHA) and Cross-Modal Knowledge Injection (CMKI). AHA performs multi-modal fusion to generate more comprehensive representations of the 3D scenario. CMKI injects the fused knowledge into the camera detection pipeline, enhancing the base model's performance. The entire model is trained in a single stage, with more details provided below.
\subsubsection{Adaptive Hierarchical Aggregation} 
\begin{figure}[t]
\centering
\includegraphics[width=1.0\columnwidth]{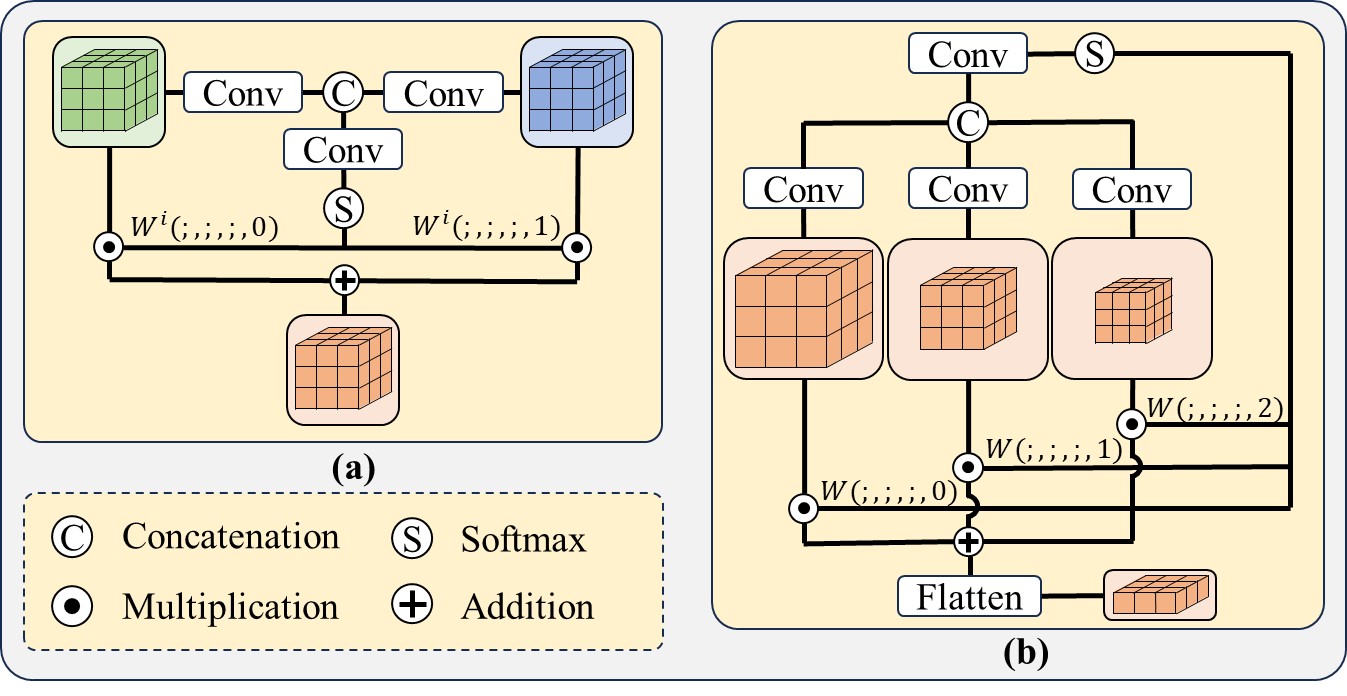}
\caption{Illustration of Adaptive Hierarchical Aggregation (AHA). LiDAR voxel features and camera pseudo voxel features are fused at different voxel scales, as shown in (a). Hierarchical fusion features are aggregated in a flexible way to obtain the output BEV features, as shown in (b).}
\label{myfig3}
\vspace{-0.5cm} 
\end{figure}

Adaptive Hierarchical Aggregation (AHA) aims to generate fine-grained fusion features by fully utilizing both the geometric information from point clouds and the semantic information from images. To achieve this, we perform multi-modal fusion at various granularities and then adaptively aggregate the multi-scale fusion features, as illustrated in Figure \ref{myfig3}. For the camera information, the camera-only base model takes multi-view images as input and projects semantic features into 3D space to obtain pseudo voxel features $F_c$. Point clouds are also voxelized to generate LiDAR voxel features $F_l$ through voxel encoding. Different voxelization scales are considered in both modalities, allowing for comprehensive geometry-semantics interaction at multiple granularities. The multi-scale pseudo voxel features from images are denoted as $F_{c}^{i}\in\mathbb{R}^{\frac{D}{2^iS_D}\times\frac{H}{2^iS_H}\times\frac{W}{2^iS_W}\times C}$ and LiDAR voxel features as $F_{l}^{i}\in\mathbb{R}^{\frac{D}{2^iS_D}\times\frac{H}{2^iS_H}\times\frac{W}{2^iS_W}\times C}$, where $D$, $H$, $W$ is the volumetric size of the entire space, $S_D$, $S_H$, $S_W$ is the minimal voxel size in three dimensions and $i = 0,1,2$. We dynamically integrate features from $F_l^i$ and $F_c^i$:
\begin{equation}\label{eq1}
\begin{aligned}
    W^i=Softmax(C_{lc}^i[C&_{l}^i(F_l^i),C_{c}^i(F_c^i)]),\\
    F_{lc}^i=W^i(:,:,:,0)\odot F_l^i+&W^i(:,:,:,1)\odot F_c^i,
\end{aligned}
\end{equation}
where $C_{lc}^i$, $C_{l}^i$, $C_{c}^i$ are 3D convolutions, $[\cdot,\cdot]$ is the concatenation along feature channel, $Softmax$ is softmax function along feature channel, $W^i\in\mathbb{R}^{\frac{D}{2^iS_D}\times\frac{H}{2^iS_H}\times\frac{W}{2^iS_W}\times 2}$ is attention weights, $\odot$ represents broadcasting multiplication.

Multi-level fusion features $F_{lc}^i\in\mathbb{R}^{\frac{D}{2^iS_D}\times\frac{H}{2^iS_H}\times\frac{W}{2^iS_W}\times C}$ have different perception ability for objects of different sizes. For example, $F_{lc}^0$ is more suitable for small object detection because it partitions the entire 3D space with the minimal voxel size and preserves more fine-grained object information. In order to combine fusion features of different voxel scales, we first adjust the grid shape of $F_{lc}^0$ and $F_{lc}^2$ to be the same as $F_{lc}^1$ through interpolation. Then we aggregate the hierarchical features as:
\begin{equation}\label{eq2}
\begin{aligned}
    W =Softmax(C_{f}[C_{f}^0(F&_{lc}^0),C_{f}^1(F_{lc}^1),C_{f}^2(F_{lc}^2)]),\\
    F_{f} =Flatten(\sum_{i=0}^{2}&W(:,:,:,i)\odot F^i_{lc}),
\end{aligned}
\end{equation}
where $C_{f}^0$, $C_{f}^1$, $C_{f}^2$, $C_{f}$ are 3D convolutions, $W\in\mathbb{R}^{\frac{D}{2S_D}\times\frac{H}{2S_H}\times\frac{W}{2S_W}\times 3}$ is attention weights, $Flatten$ is vertical dimension reduction, $F_f\in\mathbb{R}^{\frac{H}{2S_H}\times\frac{W}{2S_W}\times C}$ is the fusion BEV feature after adaptive hierarchical aggregation.
\subsubsection{Cross-Modal Knowledge Injection} Built on fusion features $F_f$, Cross-Modal Knowledge Injection (CMKI) endows camera detectors with better 3D perception ability. We first get camera-only BEV features from $F^1_c$:
\begin{equation}\label{eq3}
\begin{aligned}
    F_{c} = Flatten(F^1_c).
\end{aligned}
\end{equation}

In the camera-only base model, $F_c$ is sent to the BEV encoder $En_{BEV}$ to extract high-level semantic information of the driving scenario and the detection head $H$ outputs response features:
\begin{equation}\label{eq4}
\begin{aligned}
    F_{c}^{en} = En_{BEV}(F_{c}), F_{c}^{resp} = H(F_{c}^{en}).
\end{aligned}
\end{equation}
We find that though there is a cross-modal domain gap, the BEV encoder and detection head also work for fusion detection as long as multi-modal features $F_f$ join model training:
\begin{equation}\label{eq5}
\begin{aligned}
    F_{f}^{en} = En_{BEV}(F_{f}), F_{f}^{resp} = H(F_{f}^{en}).
\end{aligned}
\end{equation}
We take this as a hybrid supervision strategy because both kinds of BEV features share the same model parameters and are supervised by the same ground truth.

\begin{figure}[t]
\centering
\includegraphics[width=1.0\columnwidth]{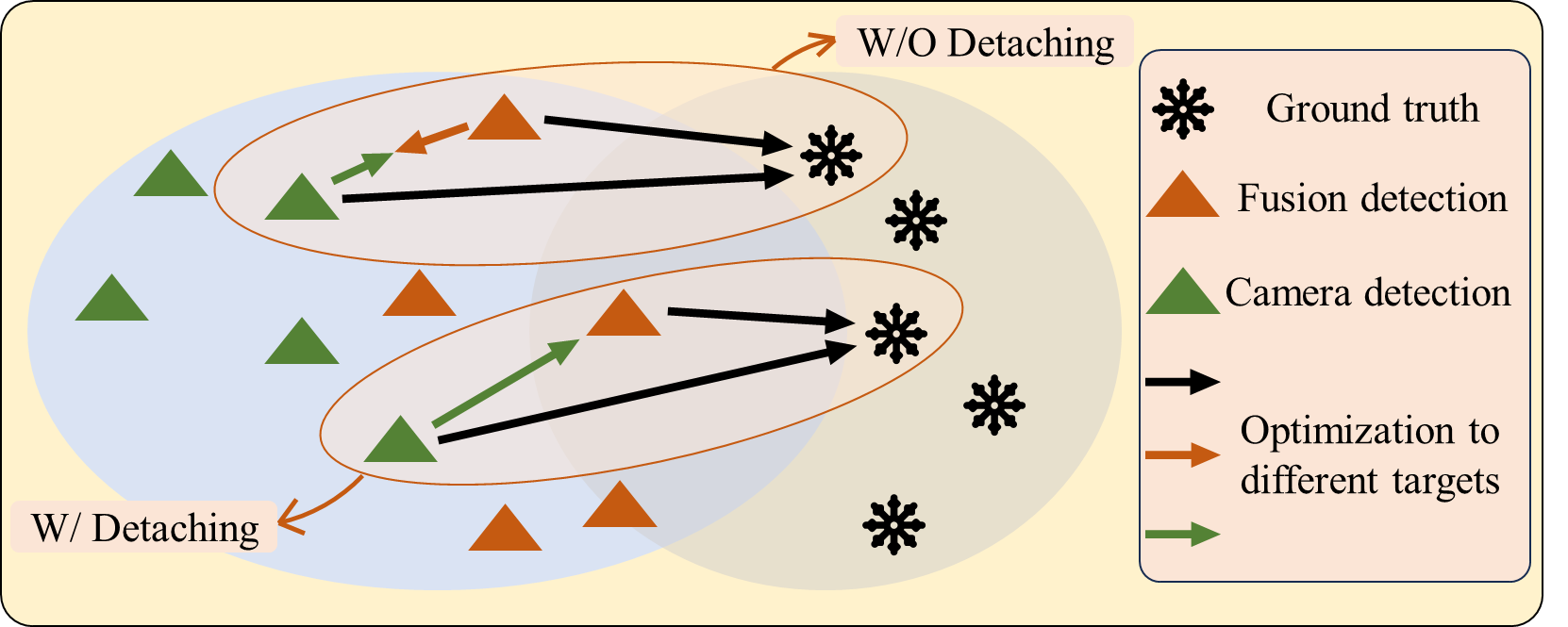}
\caption{Illustration of data distribution change during PromptDet training. LiDAR-camera detection and camera-only detection are both supervised by the ground truth. If fusion features are not detached in Cross-Modal Knowledge Injection (CMKI), they are also supervised by camera-only ones, which leads to inferior detection performance. }
\label{myfig4}
\vspace{-0.5cm} 
\end{figure}

We conduct CMKI adopting the same approach as Unidistill with some modifications. For feature knowledge injection, considering $F_f$ is derivated from $F_{c}^1$, directly calculating the feature loss between $F_f$ and $F_c$ leads to difficulty in convergence and inferior performance. Therefore, we propose the imitation module to help $F_{c}$ learn better from $F_f$ and Equation \ref{eq3} is replaced by:
\begin{equation}\label{eq6}
\begin{aligned}
    F_{c} = C_{imi}^{2D}(Flatten(C_{imi}^{3D}(F_{c}^1))),
\end{aligned}
\end{equation}
where $C_{imi}^{3D}$ is a 3D convolutional layer and $C_{imi}^{2D}$ is a 2D convolutional layer. The imitation module reserves during model inference and the added two convolutional layers almost have no impact on inference time. For relation knowledge injection and response knowledge injection, we respectively calculate the relation loss between $F_{c}^{en}$ and $F_{f}^{en}$ and the response loss between $F_{c}^{resp}$ and $F_{f}^{resp}$. To make sure the multi-modal features are not influenced by the camera-only branch, we detach the multi-modal features to stop the backward gradient during CMKI, as illustrated in Figure \ref{myfig4}. The CMKI loss is shown as:
\begin{equation}\label{eq7}
\begin{aligned}
    \mathcal{L}_{fea} = &FeaDistill(F_f.detach(),F_{c},GT),\\
    \mathcal{L}_{rel} = R&elDistill(F_f^{en}.detach(),F_{c}^{en},GT),\\
    \mathcal{L}_{resp} = Res&pDistill(F_f^{resp}.detach(),F_{c}^{resp},GT),
\end{aligned}
\end{equation}
where $.detach()$ means detached from the computational graph, $FeaDistill()$, $RelDistill()$, $RespDistill()$ are the knowledge distillation methods mentioned in Unidistill, $GT$ is the ground truth, which contains the category and location of objects.
\subsubsection{Training and Inference} During PromptDet training, the LiDAR modality switch in Figure \ref{myfig2} is turned off. Multi-modal BEV features generated by AHA are supervised by the ground truth. Camera-only BEV features are supervised by both the ground truth and fusion branch. The whole model training is completed in a single stage and the overall training objective $\mathcal{L}$ is defined as:
\begin{equation}\label{eq8}
\begin{aligned}
    \mathcal{L} = \mathcal{L}_{det}^{f}+\mathcal{L}_{det}^{c}+\lambda_1\mathcal{L}_{fea}+\lambda_2\mathcal{L}_{rel}+\lambda_3\mathcal{L}_{resp},
\end{aligned}
\end{equation}
where $\mathcal{L}_{det}^{f}$ and $\mathcal{L}_{det}^{c}$ are multi-modal detection loss and camera-only detection loss respectively, which apply the same setting as the baseline. We empirically set $\lambda_1$, $\lambda_2$, $\lambda_3$, and experiments on the sensitivity analysis are shown in supplementary.

There are two situations during PromptDet inference. When images and LiDAR points are both available, the LiDAR modality switch is turned off and PromptDet conducts multi-modal detection. When there are only images for inference, the LiDAR modality switch is turned on and PromptDet degenerates into the camera-only base model but achieves better performance. 
\label{promptDet}
\subsection{Properties of PromptDet} 
As a novel 3D object detection framework, PromptDet offers three main advantages: (1) Integration of LiDAR Modality: PromptDet introduces LiDAR data to the camera-based detector, resulting in significant performance gains with only a minimal increase in parameters—about 1\% of the base model's total. This increase primarily comes from convolutional layers in the Adaptive Hierarchical Aggregation and the imitation module.
(2) Enhanced Perception Through Cross-Modal Knowledge Injection: PromptDet enhances the camera detector's perception capabilities via online cross-modal knowledge transfer. This approach is easy and simple during model training.
(3) Flexibility of LiDAR Modality During Inference: The LiDAR modality in PromptDet is flexible during inference, thanks to an adjustable LiDAR modality switch. This feature ensures that PromptDet remains a robust 3D object detector even in cases of LiDAR malfunction.
\label{benefits}

\section{Experiment}
\subsection{Experimental Setup}
\subsubsection{Dataset and Metrics}
We evaluate our framework on the nuScenes dataset \cite{nuscenes}, one of the most challenging benchmarks in autonomous driving. The dataset consists of 1,000 driving scenarios, divided into 700 for training, 150 for validation, and 150 for testing. Each scenario includes synchronized LiDAR point clouds and images from six surrounding cameras, enabling comprehensive multi-modal perception. In alignment with the official evaluation protocol, we report the nuScenes Detection Score (NDS), mean Average Precision (mAP), and five additional metrics across ten popular classes.
\begin{table*}[!t]
  \centering
  \fontsize{8.5}{9.5}\selectfont
  \renewcommand{\arraystretch}{1.2}
    \begin{tabular}{c|cc|cc|ccccc}
    \Xhline{1.0pt}
    Method & Mode & Backbone & mAP$\uparrow$ & NDS$\uparrow$ & mATE$\downarrow$ & mASE$\downarrow$ & mAOE$\downarrow$ & mAVE$\downarrow$ & mAAE$\downarrow$ \\
    \Xhline{1.0pt}
    BEVFusion \cite{bevfusion_beida} & L\&C & VoxelNet SwinT & 67.9 & 71.0 & - & - & - & - & - \\
    BEVFusion* \cite{bevfusion_beida} & L\&C & PointPilliars SwinT & 53.5 & 60.4 & - & - & - & - & - \\
    PointPilliars \cite{pointpillars} & L & PointPilliars & 45.3 & 30.5 & - & - & - & - & - \\
    \hline
    FCOS3D \cite{fcos3d} & C & R101 & 34.3 & 41.5 & 72.5 & 26.3 & 42.2 & 129.2 & 15.3\\
    PETR \cite{petr} & C & R101 & 35.7 & 42.1 & 71.0 & 27.0 & 49.0 & 88.5 & 22.4\\
    PETR & C & SwinT & 36.1 & 43.1 & 73.2 & 27.3 & 49.7 & 80.8 & 18.5\\
    DETR3D \cite{detr3d} & C & R101 & 34.9 & 43.4 & 71.6 & 26.8 & 37.9 & 84.2 & 20.0\\
    BEVFusion-C \cite{bevfusion_mit} & C & SwinT & 35.6 & 41.2 & 66.8 & 27.3 & 56.1 & 89.6 & 25.9\\
    BEVFormer \cite{bevformer} & C & R101 & 41.6 & 51.7 & 67.3 & 27.4 & 37.2 & 39.4 & 19.8\\
    BEVDepth \cite{bevdepth} & C & R101 & 41.2 & 53.5 & - & -  & -  &  - & - \\
    \hline
    \multirow{2}{*}{PromptDet} & C & R101 & \textcolor{red}{43.3} & \textcolor{red}{56.9} & \textcolor{red}{44.5} & \textcolor{red}{23.6} & \textcolor{red}{35.1} &  \textcolor{red}{32.1} & \textcolor{red}{12.6}\\
     & L\&C & R101 & \textcolor{blue}{56.2} & \textcolor{blue}{64.1} &  \textcolor{blue}{39.8} & \textcolor{blue}{24.1} & \textcolor{blue}{34.6} & \textcolor{blue}{29.5} & \textcolor{blue}{12.1}\\
    \Xhline{1.0pt}
    \end{tabular}
  \caption{Performance comparison on the nuScenes validation set. L and C represent the input modality, \textit{i.e.}, LiDAR and camera. * means the PointPillars version of BEVFusion.}
  \label{mytab1}
  \vspace{-0.5cm} 
\end{table*}
\begin{table}[!t]
  \centering
  \fontsize{8.5}{9.5}\selectfont
  \renewcommand{\arraystretch}{1.1}
    \begin{tabular}{c|cc|cc}
    \Xhline{1.0pt}
    Method & LaP & Mode & mAP & NDS \\
    \Xhline{1.0pt}
     \multirow{3}{*}{BEVDet} & - & C & 27.7 & 35.3 \\
     & \checkmark & L\&C & 50.5 & 56.4 \\
     & \checkmark & C & 30.1 & 39.3 \\
    \hline
    \multirow{3}{*}{BEVDet4D} & - & C & 28.9 & 41.4 \\
     & \checkmark & L\&C & 52.3 & 58.5 \\
     & \checkmark & C & 30.7 & 44.5 \\
    \hline
    \multirow{3}{*}{BEVDepth} & - & C & 32.8 & 44.1 \\
     & \checkmark & L\&C & 55.9 & 62.7 \\
     & \checkmark & C & 35.2 & 48.0 \\
    \Xhline{1.0pt}
    \end{tabular}
  \caption{Performance comparison of different camera detectors with our proposed LiDAR-assisted Prompter, which is shortened as LaP.}
  \label{mytab2}
  \vspace{-0.5cm} 
\end{table}
\subsubsection{Implementation Details}
We use three camera-only BEVDet-series detectors—BEVDet, BEVDepth, and BEVDet4D—as the base models to evaluate the generalization of our method. All experiments are conducted in PyTorch using 4 NVIDIA A40 GPUs (45GB memory), based on the MMDetection3D \cite{mmdetection3d} codebase. Unless otherwise specified, we use ResNet-50 \cite{resnet} as the image backbone, with images resized to 256×704. The detection range is set to [-51.2m, 51.2m] × [-51.2m, 51.2m], and we use dynamic point cloud voxelization \cite{dynamic} to reduce unnecessary computational costs. The moderate voxel size is set to [0.8m, 0.8m, 0.8m]. AdamW \cite{adamw} is used as the optimizer with a step-scheduled learning rate. When compared with other methods, we train the model for 30 epochs with CBGS \cite{cbgs}; other experiments are trained for 36 epochs without CBGS. We apply synchronous BEV data augmentations to both LiDAR and camera modalities for better generalization. For BEVDet4D, we perform temporal camera fusion at the voxel level.

\subsection{Main Results}
We begin by using BEVDepth as the base model to compare our method, PromptDet, with state-of-the-art methods on the nuScenes validation set. The results are shown in Table \ref{mytab1}. Using ResNet-101 \cite{resnet} as the image backbone, our method, PromptDet-C (with images only), improves the baseline by 2.1\% mAP and 3.4\% NDS. When point clouds are available, the lightweight multi-modal detector, PromptDet-LC, outperforms the PointPillars version of BEVFusion \cite{bevfusion_beida} by 2.7\% mAP and 3.7\% NDS, without using the time-consuming LiDAR backbone. Next, we compare detection performance before and after introducing the LiDAR-assisted prompter into three representative camera-based models to evaluate the generalization of our method. The experiments, conducted on the nuScenes validation set, are summarized in Table \ref{mytab2}. We observe significant performance gains when point clouds are available, with both mAP and NDS improving substantially for all three baselines. The largest performance margin (\textit{i.e.}, 22.8\% mAP and 21.1\% NDS) is achieved by adding the LiDAR-assisted prompter to BEVDet with few additional parameters compared with the base model. Even when only images are used for inference, PromptDet still enhances these camera-based models. For the base model BEVDet, the most significant improvement is 2.4\% mAP and 4.0\% NDS. For the temporal fusion model BEVDet4D, we still see gains of 1.8\% mAP and 3.1\% NDS.
\subsection{Ablation Studies}
We conduct the ablations on the detailed components of PromptDet with BEVDet as the base model.
\begin{table}[!t]
  \centering
  \fontsize{8.5}{9.5}\selectfont
  \renewcommand{\arraystretch}{1.1}
    \begin{tabular}{c|ccc|cc|cc}
    \Xhline{1.0pt}
    & AHA & CMKI & Det. & mAP & NDS & mAP* & NDS*\\
    \Xhline{1.0pt}
    a & & & & 27.1 & 34.8 & 49.7 & 55.1\\
     b & \checkmark & & & 27.3 & 34.2 & 50.9 & 55.9  \\
     c & & \checkmark & & 29.1 & 38.1 & 48.8 & 54.3 \\
     d & & \checkmark & \checkmark & 29.6 & 39.0 & 49.8 & 55.1\\
     e & \checkmark & \checkmark & & 29.3 & 38.9 & 49.4 & 56.0 \\
     f & \checkmark & \checkmark & \checkmark & 30.1 & 39.3 & 50.5 & 56.4 \\
    \Xhline{1.0pt}
    \end{tabular}
  \caption{Ablation study of the model architecture and training setting. Det. means detaching fusion features in Cross-Modal Knowledge Injection (CMKI). * means multi-modal detection results of PromptDet-LC.}
  \label{mytab3}
  \vspace{-0.5cm} 
\end{table}
\subsubsection{Effect of Individual Component}
\begin{figure}[t]
\centering
\includegraphics[width=1.0\columnwidth]{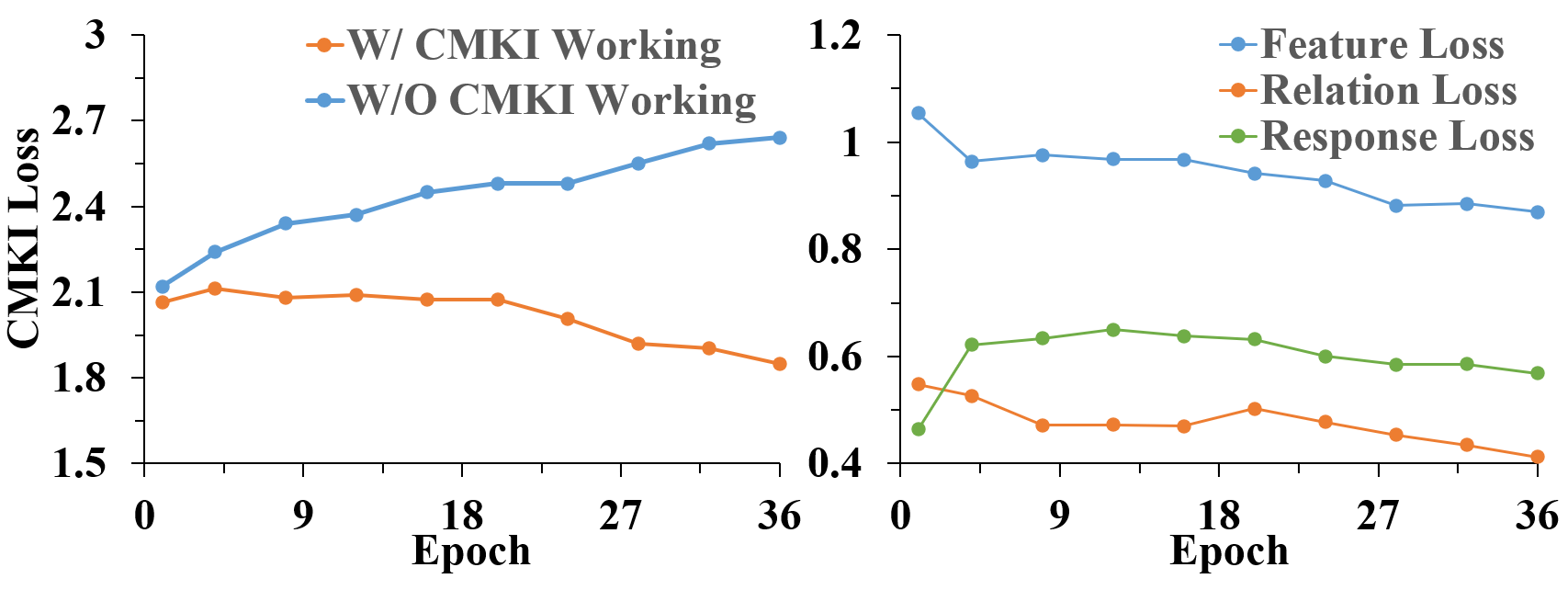}
\caption{Illustration to show the effect of Cross-Modal Knowledge Injection (CMKI). With CMKI working, the data distribution difference between fusion features and camera features narrows down and all three kinds of CMKI loss decrease slowly.}
\label{myfig5}
\end{figure}

We ablate the contribution of each component in PromptDet, as shown in Table \ref{mytab3}. In baseline (a), we fuse multi-modal features at a single scale and use hybrid supervision without CMKI. In models (c) and (d), we introduce CMKI and compare the performance with and without detaching fusion features in CMKI. CMKI improves PromptDet-C by 2.0\% mAP and 3.3\% NDS but reduces PromptDet-LC by 0.9\% mAP and 0.8\% NDS. The detachment operation reduces this performance drop, supporting our hypothesis. Finally, we introduce AHA and perform multi-modal fusion at different scales, as shown in (b), (e), and (f). Compared to ``without AHA'', PromptDet-LC shows consistent improvement. The most significant gains are in model (f), with 3.0\% mAP and 4.5\% NDS for PromptDet-C, and 0.8\% mAP and 1.3\% NDS for PromptDet-LC. To better understand the impact of CMKI during training, we show the trends of CMKI loss for models (a) and (d) as well as the detailed CMKI loss for model (d) in Figure \ref{myfig5}. Without CMKI's constraints, the loss tends to diverge during training and the data distribution difference between PromptDet-LC and PromptDet-C increases. Even with CMKI active, the loss decreases slowly, likely because PromptDet-LC and PromptDet-C are trained together to align with the ground truth. CMKI keeps the two data distributions closer and aids in cross-modal knowledge transfer.
\begin{table}[!t]
  \centering
  \fontsize{8.5}{9.5}\selectfont
  \renewcommand{\arraystretch}{1.1}
    \begin{tabular}{c|cc|cc|cc}
    \Xhline{1.0pt}
    & Conv3D & Conv2D & mAP & NDS & mAP* & NDS*\\
    \Xhline{1.0pt}
    a & & & 28.4 & 36.8 & 50.4 & 56.5\\
     b & \checkmark & & 29.6 & 39.0 & 50.1 & 56.2  \\
     c & & \checkmark & 28.9 & 37.5 & 50.1 & 56.7\\
     d & \checkmark & \checkmark & 30.1 & 39.3 & 50.5 & 56.4 \\
    \Xhline{1.0pt}
    \end{tabular}
  \caption{Ablation study of different imitation module designs. * means multi-modal detection results of PromptDet-LC.}
  \label{mytab4}
  \vspace{-0.5cm} 
\end{table}
\subsubsection{Design of Imitation Module}
The imitation module is designed to force camera-only features to learn from fusion features. It consists of a 3D convolutional layer, a flatten operation, and a 2D convolutional layer. We tested different designs of the imitation module (\textit{i.e.}, without any convolutional layers, with only a 3D convolutional layer, with only a 2D convolutional layer, and the complete architecture). As shown in Table \ref{mytab4}, different imitation modules have little impact on PromptDet-LC, but omitting all convolutional layers leads to poorer performance for PromptDet-C. Both the 2D and 3D convolutional layers help camera-only features better represent the 3D scene, with the complete design providing the most significant improvement of 1.7\% mAP and 2.5\% NDS.
\subsection{Further Discussion}
\subsubsection{Model Efficiency}
In Table \ref{mytab5}, we report the model efficiency, including parameters and inference speed. PromptDet-C adds only the imitation module to the baseline, consisting of two convolutional layers, which has minimal impact on inference time. Therefore, we focus on PromptDet-LC when point clouds are available, with all inferences conducted on a single NVIDIA A40 GPU. The LiDAR-assisted prompter adds fewer than 2\% extra parameters during training and inference. PromptDet-LC, based on BEVDepth, takes 115.4 ms for inference, running much faster than the PointPillars version of BEVFusion \cite{bevfusion_beida} while achieving better detection performance.
\subsubsection{Qualitative Results}
In Figure \ref{myfig6}, we visualize the prediction results from both the multi-camera view and LiDAR top view. Compared to BEVDet as the baseline, PromptDet-LC significantly improves perception and detects objects missed by the baseline. Even without LiDAR input, PromptDet-C reduces false positives and better localizes objects, thanks to our Cross-Modal Knowledge Injection during training.
\begin{table}[t]
  \centering
  \fontsize{8.5}{9.5}\selectfont
  \renewcommand{\arraystretch}{1.1}
    \begin{tabular}{cc|cc}
    \Xhline{1.0pt}
    Methods & Mode & Params(M) & Latency(ms)\\
    \Xhline{1.0pt}
    BEVFusion & L\&C & 90.19 & 1453.8 \\
     BEVFusion* & L\&C & 89.46 & 1428.6 \\
     \hline
     BEVDet & C & 44.25 & 41.9 \\
     +LaP & L\&C & 44.86(+1.38\%) & 100.5 \\
     \hline
     BEVDet4D & C & 52.48 & 73.0 \\
     +LaP & L\&C & 53.49(+1.92\%) & 158.4 \\
     \hline
     BEVDepth & C & 66.67 & 55.7 \\
     +LaP & L\&C & 67.28(+0.91\%) & 115.4 \\
    \Xhline{1.0pt}
    \end{tabular}
  \caption{Comparison of model efficiency. * means the PointPillars version of BEVFusion. LaP denotes our proposed LiDAR-assisted Prompter.}
  \label{mytab5}
\end{table}
\begin{figure}[t]
\centering
\includegraphics[width=1.0\columnwidth]{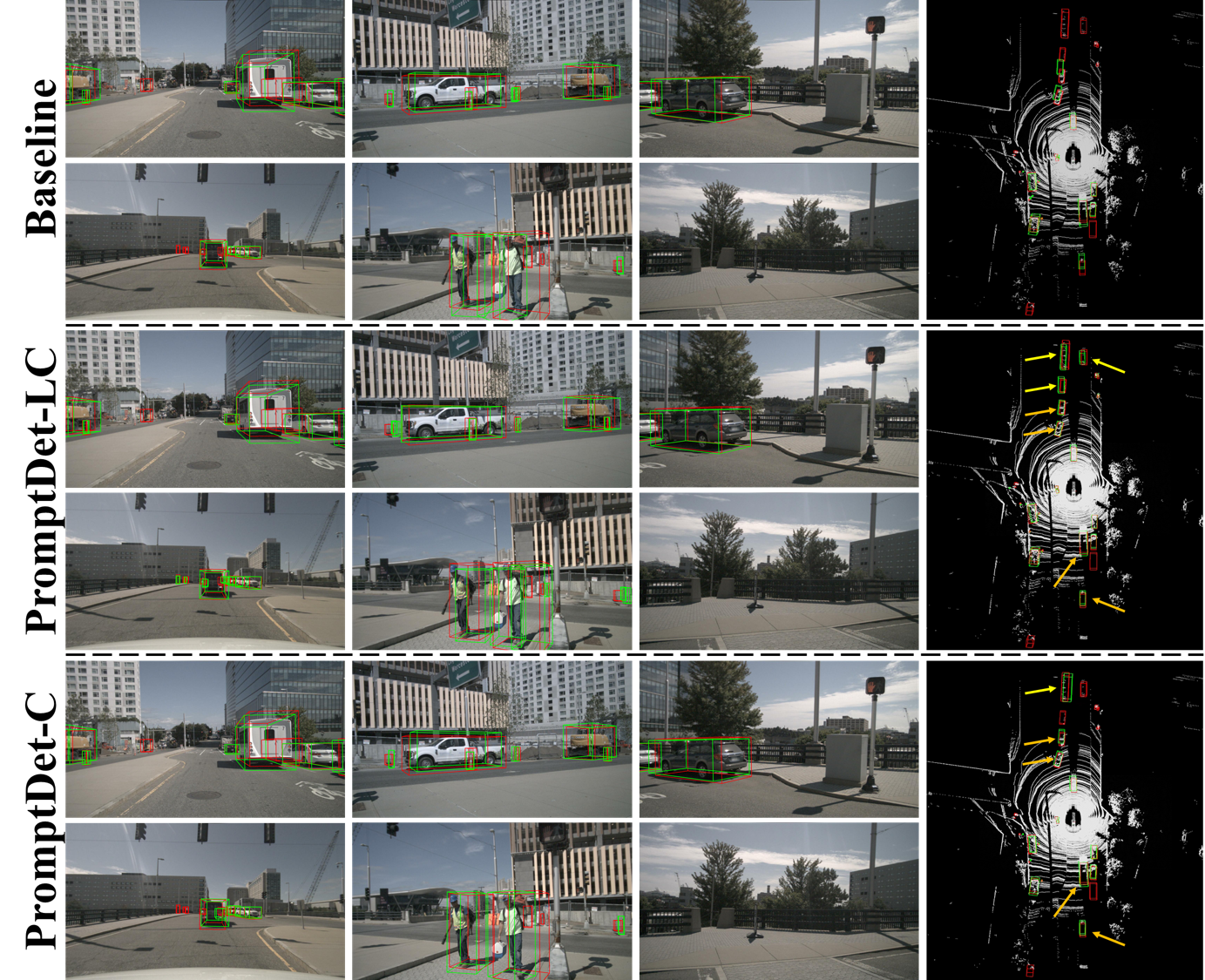}
\caption{The comparison of BEVDet \cite{bevdet} and our proposed PromptDet. The red boxes and green boxes denote the ground truth and detection results, respectively. In the LiDAR top view, We use the yellow arrows to mark the objects neglected by baseline and orange arrows to mark better localization in PromptDet-LC and PromptDet-C. }
\label{myfig6}
\vspace{-0.5cm} 
\end{figure}
\section{Conclusion}
In this paper, we propose a lightweight 3D object detection framework dubbed PromptDet, which consists of a camera detector and the LiDAR-assisted prompter. PromptDet conducts LiDAR-camera fusion through AHA and is a lightweight multi-modal detector if both images and LiDAR points are available. Thanks to CMKI, PromptDet still outperforms the baseline with only images as input. AHA and CMKI constitute the plug-and-play LiDAR-assisted prompter and the whole framework training is simple with few extra model parameters. Extensive experiments on the challenging nuScenes benchmark validate the effectiveness of our method. We believe the paradigm of PromptDet is compatible with diverse camera-based detectors and can be extended to more multi-camera perception tasks, such as occupancy prediction, BEV segmentation, and so on.

\section*{Acknowledgments}
This work was supported by USTC-JAC Smart Electric Vehicle Joint Lab, and by the Key Science and Technology Program of Anhui under Grant 202203a05020012.

\appendix
\section{Appendices}
\subsection{Additional Experiments}
\subsubsection{Sensitivity Analysis on Hyper-parameters}
For different base models, we adjust the hyper-parameters, $\lambda_1$, $\lambda_2$ and $\lambda_3$, to balance the CMKI loss and the detection loss. Table \ref{mytab7} shows the combinations of hyper-parameter settings and base models. (a), (b), and (c) are the default settings for BEVDet, BEVDet4D, and BEVDepth, respectively. Despite using different hyper-parameters, all base models show performance gains. For BEVDet and BEVDet4D, settings (a) and (b) yield better results. For BEVDepth, setting (c) is more suitable since the CMKI loss needs to be increased to balance the added depth estimation loss.
\subsubsection{Other Experiments}
In Table \ref{mytab8}, we retrain PromptDet with only LiDAR-camera fusion, referred to as Prompt-FL. Prompt-FL achieves slightly better performance than PromptDet without camera branch training. However, Prompt-FL doesn't adapt to camera-only detection. In contrast, our PromptDet offers a much better balance between fusion detection and camera detection. In Table \ref{mytab9}, we replace the convolutional layers with the 3D deformable attention in AHA. Both of them can be used to conduct multi-modal fusion. For 3D deformable attention, we choose the fusion voxel features with the medium voxel size as the query and fusion voxel features of all three different sizes as the key and value. Though there are fewer parameters with deformable attention, the performance decreases a little. Therefore, we utilize the convolution-based method as the default setting. 
\subsection{Qualitative Results}
\subsubsection{Prediction Results Visualization}
In Figure \ref{myfig6}, we visualize the prediction results from both the multi-camera view and the LiDAR top view. Compared to the baseline, BEVDet, PromptDet-LC can significantly improve perception and detect objects missed by the baseline. Even without LiDAR input, PromptDet-C can reduce false positives and better localize objects, thanks to our Cross-Modal Knowledge Injection during training.

\subsubsection{Feature Map Visualization}
\begin{figure*}[t]
\centering
\includegraphics[width=1.0\textwidth]{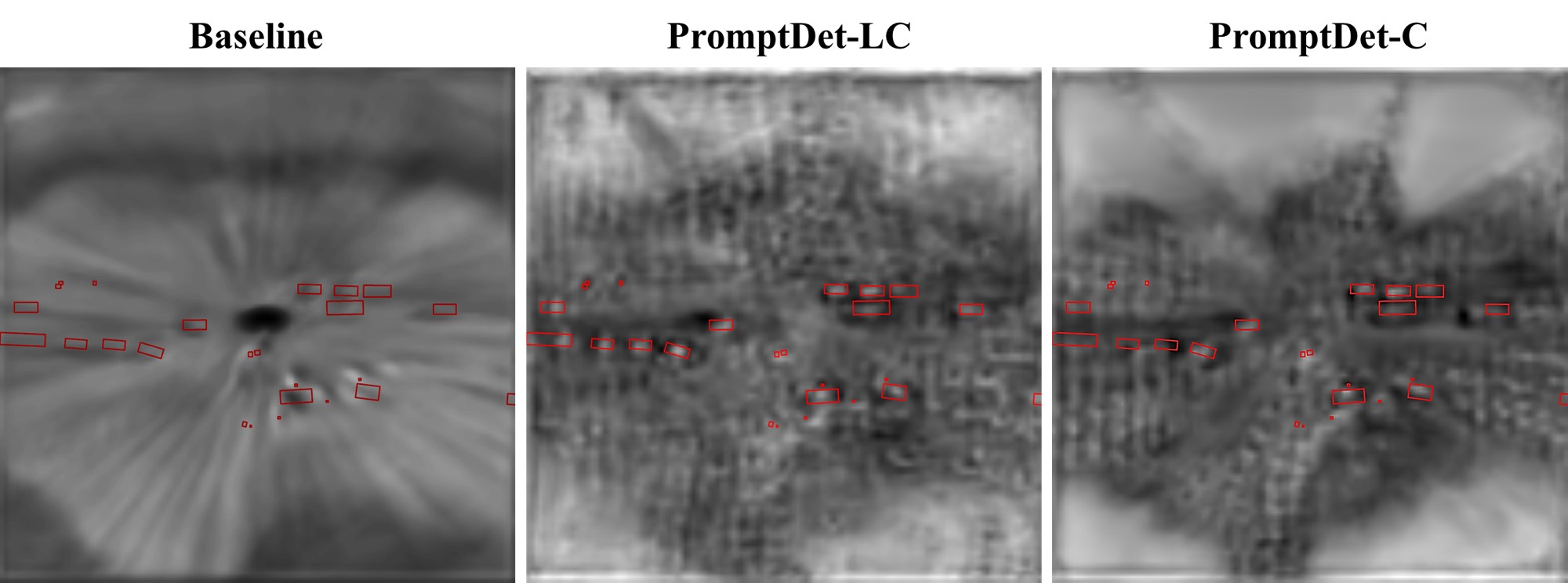}
\caption{Visualization of feature maps in baseline and our proposed PromptDet. The red boxes denote the ground truth in the LiDAR top view.}
\label{myfig7}
\end{figure*}

As shown in Figure \ref{myfig7}, we compare BEV feature maps generated by the BEV encoder from different detectors. We calculate the mean across the channel dimension to view all feature maps. The BEV features from the baseline BEVDet are radial, which is due to the lack of precise depth localization information. PromptDet-LC has much clearer object boundaries because LiDAR points provide precise geometric information for better localization and the BEV features are granular. PromptDet-C successfully learns the granular data distribution from PromptDet-LC, producing sharper feature maps than the baseline, especially in some regions near objects.
\begin{table}[t]
  \centering
  \fontsize{7}{8}\selectfont
  \renewcommand{\arraystretch}{1.0}
    \begin{tabular}{c|c|cc|cc|cc}
    \Xhline{1.0pt}
     & \multirow{2}{*}{$\lambda_1$/$\lambda_2$/$\lambda_3$} & \multicolumn{2}{c|}{BEVDet} & \multicolumn{2}{c|}{BEVDet4D} & \multicolumn{2}{c}{BEVDepth} \\
     & & mAP & NDS & mAP & NDS & mAP & NDS \\
     \Xhline{1.0pt}
     a & 1.1/8.0/2.0 & 30.1 & 39.3 & 30.9 & 44.1 & 33.9 & 46.5\\
     b & 1.5/10.0/2.5 & 29.5 & 39.2 & 30.7 & 44.5 & 34.3 & 45.7\\
     c & 8.0/25.0/10.0 & 28.5 & 37.1 & 29.4 & 43.5 & 35.2 & 48.0\\
    \Xhline{1.0pt}
    \end{tabular}
  \caption{Performance comparison of PromptDet-C with different hyper-parameters.}
  \label{mytab7}
\end{table}
\begin{table}[t]
  \centering
  \fontsize{8.5}{9.5}\selectfont
  \renewcommand{\arraystretch}{1.0}
    \begin{tabular}{c|c|cc|cc}
    \Xhline{1.0pt}
    & Method & mAP & NDS & mAP* & NDS*\\
    \Xhline{1.0pt}
    a & PromptDet-FL& - & - & 52.8 & 58.9\\
     b & PromptDet & 30.1 & 39.3 & 50.5 & 56.4  \\
    \Xhline{1.0pt}
    \end{tabular}
  \caption{Performance comparison of PromptDet-FL and PromptDet. We take BEVDet as the base model. * means multi-modal detection results.}
  \label{mytab8}
\end{table}
\begin{table}[!]
  \centering
  \fontsize{8.5}{9.5}\selectfont
  \renewcommand{\arraystretch}{1.0}
    \begin{tabular}{c|c|cc|cc|c}
    \Xhline{1.0pt}
    & Method & mAP & NDS & Params(M)\\
    \Xhline{1.0pt}
    a & Conv3D& 50.5 & 56.4 & 44.86\\
     b & DeformAttn3D & 49.7 & 56.0 & 44.57 \\
    \Xhline{1.0pt}
    \end{tabular}
  \caption{Performance comparison of using convolutional layers and 3D deformable attention in AHA. We take BEVDet as the base model and report the performance of PromptDet-LC here.}
  \label{mytab9}
\end{table}
\begin{figure*}[t]
\centering
\includegraphics[width=1.0\textwidth]{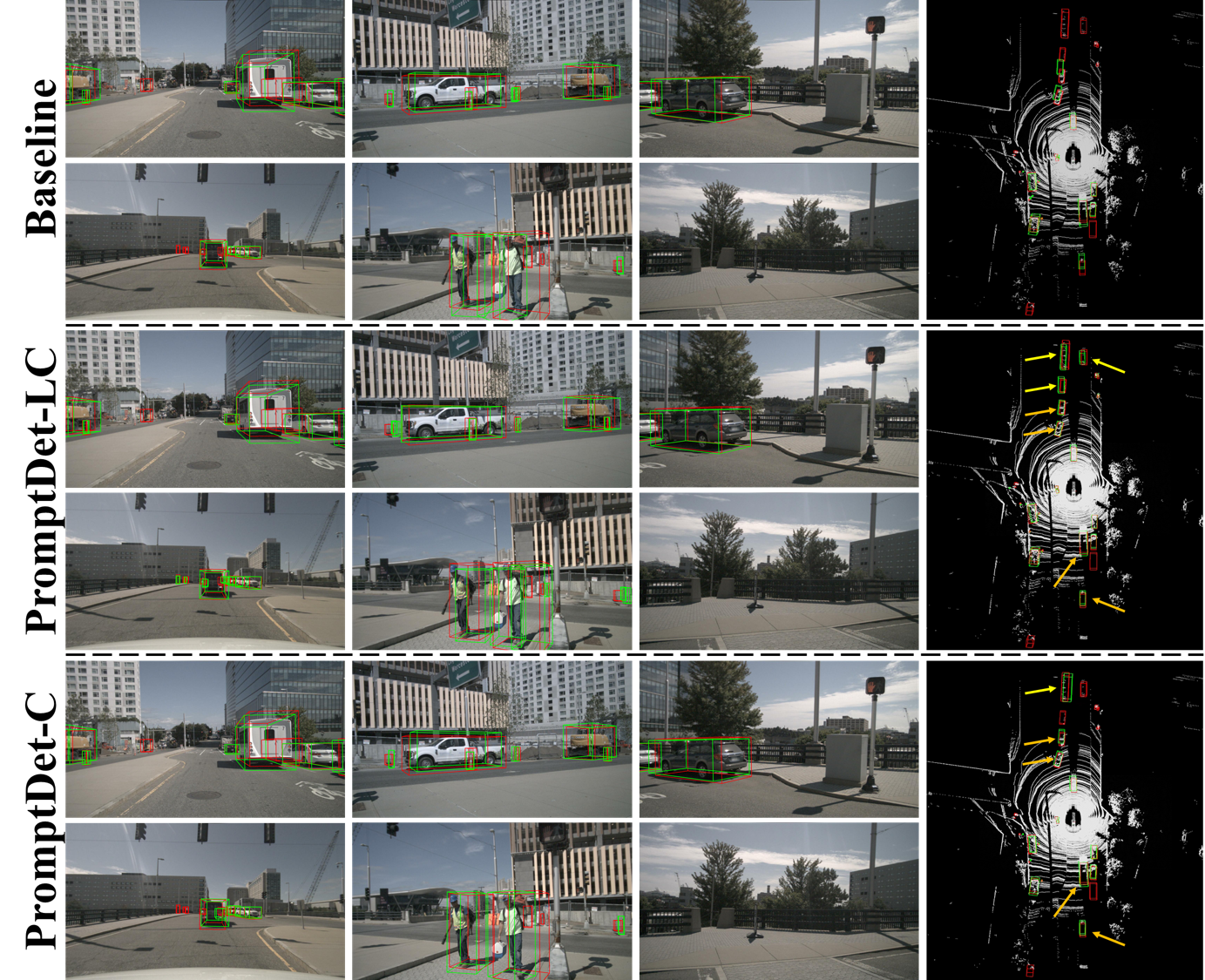}
\caption{Comparison of the baseline BEVDet and our proposed PromptDet. The red boxes and green boxes denote the ground truth and detection results, respectively. In the LiDAR top view, We use the yellow arrows to mark the objects neglected by baseline and orange arrows to mark better localization in PromptDet-LC and PromptDet-C. }
\label{myfig6}
\end{figure*}

\bibliography{aaai25}

\end{document}